\documentclass[runningheads]{llncs}
\usepackage{graphicx}

\usepackage{xcolor}
\usepackage{multirow}
\usepackage{amsmath}

\usepackage{soul}

\def\PaperID#1{\gdef\@PaperID{#1}}
\gdef\@PaperID{}
\newif\ifreview
\reviewtrue

\ifreview
\newcommand{\rv}[2][]{%
    \if\relax\detokenize{#1}\relax%
        \textcolor{blue}{#2}%
    \else%
        \textcolor{gray}{\st{#1}}\textcolor{red}{#2}%
    \fi%
}
\else
\newcommand{\rv}[2][]{#2}
\fi
\newif\iffinalversion
\def\finalversiontext#1%
   {%
   \iffinalversion%
   #1%
   \else%
   [this content was suppressed due to blind submission]%
   \fi%
   }

\PaperID{\#89}
\finalversiontrue

\begin{document}
%
\thispagestyle{empty}
{\noindent\Large Springer Copyright Notice}\\[1pt]

{\noindent Copyright (c) 2021 Springer

\noindent This work is subject to copyright. All rights are reserved by the Publisher, whether the whole or part of the material is concerned, specifically the rights of translation, reprinting, reuse of illustrations, recitation,broadcasting, reproduction on microfilms or in
any other physical way, and transmission or information storage and retrieval, electronic adaptation, computer software, or by similar or dissimilar methodology now known or hereafter developed.}\\[1em]

{\noindent\large Accepted to be published in: 2021 25th Iberoamerican Congress on Pattern Recognition (CIARP'21), May 10--13, 2021.}\\[1in]

{\noindent Cite as:}\\[1pt]

{\setlength{\fboxrule}{1pt}
 \fbox{\parbox{0.95\textwidth}{S. F. dos Santos and J. Almeida, ``Less is More: Accelerating Faster Neural Networks Straight from JPEG,'' in \emph{2021 25th Iberoamerican Congress on Pattern Recognition (CIARP)}, Porto, Portugal, 2021, pp. 237--247, doi: 10.1007/978-3-030-93420-0\_23}}}\\[1in]

{\noindent BibTeX:}\\[1pt]

{\setlength{\fboxrule}{1pt}
 \fbox{\parbox{1.1\textwidth}{
 @InProceedings\{CIARP\_2021\_Santos,

 \begin{tabular}{lll}
  & author    & = \{S. F. \{dos Santos\} and
               J. \{Almeida\}\},\\

  & title     & = \{Less is More: Accelerating Faster Neural Networks Straight from JPEG\},\\

  & pages     & = \{237--247\},\\

  & booktitle & = \{2021 25th Iberoamerican Congress on Pattern Recognition (CIARP)\},\\

  & address   & = \{Porto, Portugal\},\\

  & month     & = \{May 10--13\},\\

  & year      & = \{2021\},\\

  & publisher & = \{\{Springer\}\},\\

  & doi       & = \{10.1007/978-3-030-93420-0\_23\},\\
  \end{tabular}

\}
 }}}

\clearpage

\title{Less is More: Accelerating Faster Neural Networks Straight from JPEG}
%
%
\author{
\iffinalversion
Samuel Felipe dos Santos\orcidID{0000-0001-6061-5582} \and \\
Jurandy Almeida\orcidID{0000-0002-4998-6996}
\else
Paper \@PaperID
\fi
}
\authorrunning{
\iffinalversion
S. F. Santos and J. Almeida
\fi
}
%
\institute{
\iffinalversion
Instituto de Ci\^{e}ncia e Tecnologia, Universidade Federal de S\~{a}o Paulo -- UNIFESP\\
12247-014, S\~{a}o Jos\'{e} dos Campos, SP -- Brazil\\
\email{\{felipe.samuel, jurandy.almeida\}@unifesp.br} 
\fi
}
\maketitle              
\begin{abstract}
Most image data available are often stored in a compressed format, from which JPEG is the most widespread. 
To feed this data on a convolutional neural network (CNN), a preliminary decoding process is required to obtain RGB pixels, demanding a high computational load and memory usage.
For this reason, the design of CNNs for processing JPEG compressed data has gained attention in recent years.
In most existing works, typical CNN architectures are adapted to facilitate the learning with the DCT coefficients rather than RGB pixels.
Although they are effective, their architectural changes either raise the computational costs or neglect relevant information from DCT inputs.
In this paper, we examine different ways of speeding up CNNs designed for DCT inputs , exploiting learning strategies to reduce the computational complexity by taking full advantage of DCT inputs.
Our experiments were conducted on the ImageNet dataset.
Results show that learning how to combine all DCT inputs in a data-driven fashion is better than discarding them by hand, and its combination with a reduction of layers has proven to be effective for reducing the computational costs while retaining accuracy.
\keywords{Deep Learning  \and Convolutional Neural Networks \and JPEG \and Discrete Cosine Transform \and Frequency Domain}
\end{abstract}
%
%
%

\section{Introduction}
Convolutional neural networks (CNNs) have achieved state-of-the-art performance in several computer vision tasks, such as, classification, segmentation, object detection, image super resolution, denoising, medical images, autonomous driving, road surveillance, among others~\cite{ITSC_2019_Deguerre,li2019learning}.
However, in order to achieve this performance, increasingly deeper architectures have been used, making computational cost one of the main problems faced by deep learning models~\cite{ICCV_2019_Ehrlich}.

For storage and transmission purposes, most image data available are often stored in a compressed format, like JPEG, PNG and GIF~\cite{ITSC_2019_Deguerre}.
To use this data with a typical CNN, it would be required to decode it to obtain RGB images, a task demanding high memory and computational cost~\cite{ITSC_2019_Deguerre}.
A possible alternative is to design CNNs capable of learning with DCT coefficients rather than RGB pixels~\cite{ITSC_2019_Deguerre,ICCV_2019_Ehrlich,NIPS_2018_Gueguen,ICIP_2020_Santos}.
These coefficients can be easily extracted by partial decoding JPEG compressed data, saving computational cost. 

In this paper, we investigate strategies to accelerate CNNs designed for the JPEG compressed domain. 
The starting point for our study is a state-of-the-art CNN proposed by Gueguen~et~al.~\cite{NIPS_2018_Gueguen}, which is a modified version of the ResNet-50~\cite{CVPR_2016_He}.
However, the changes introduced by Gueguen~et~al.~\cite{NIPS_2018_Gueguen} in the ResNet-50 raised its computational complexity and number of parameters.
To alleviate these drawbacks, Santos~et~al.~\cite{ICIP_2020_Santos} proposed to feed the network with the lowest frequency DCT coefficients, thus losing image details irretrievably.
Here, we explore smart strategies to reduce the network computation complexity without sacrificing rich information provided by the DCT coefficients.

Experiments were conducted on the ImageNet dataset, both in a subset and in the whole.
Our reported results indicate that learning how to combine all DCT inputs in a data-driven fashion is better than discarding them by hand.
We also found that skipping some stages of the network is beneficial, decreasing its computational costs, while also increasing the performance. 

The remainder of this paper is organized as follows. 
Section~\ref{sec:jpeg} briefly reviews the JPEG standard.
Section~\ref{sec:related} discusses related work.
Section~\ref{sec:approach} describes strategies to speed-up CNNs designed for DCT input.
Section~\ref{sec:experiments} presents the experimental setup and reports our results.
Finally, Section~\ref{sec:conclusions} offer our conclusions.

\section{JPEG compression}
\label{sec:jpeg}
The JPEG standard (ISO/IEC 10918) was created in 1992 and is currently the most widely-used image coding technology for lossy compression of digital images. 
The basic steps of the JPEG compression algorithm are described as follows.
Initially, the representation of the colors in the image is converted from RGB to YCbCr, which is composed of one luminance component (Y), representing the brightness, and two chrominance components, Cb and Cr, representing the color. 
Also, the Cb and Cr components are down-sampled horizontally and vertically by a factor of 2, for human vision is more sensitive to brightness details than to color details.
Then, each of the three components is partitioned into blocks of 8x8 pixels and 128 is subtracted from all the pixel values. 
Next, each block is converted to a frequency domain representation by the forward discrete cosine transform~(DCT).
The result is an 8x8 block of frequency coefficient values, each corresponding to the respective DCT basis functions, with the zero-frequency coefficient (DC term) in the upper left and increasing in frequency to the right and down.
The amplitudes of the frequency coefficients are quantized by dividing each coefficient by a respective quantization value defined in quantization tables, followed by rounding the result to the the nearest integer.
High-frequency coefficients are approximated more coarsely than low-frequency coefficients, for human vision is fairly good at seeing small variations in color or brightness over large areas, but fails to distinguish the exact strength of high-frequency brightness variations.
The quality setting of the encoder affects the extent to which the resolution of each frequency component is reduced.
If an excessively low-quality setting is used, most high-frequency coefficients are reduced to zero and thus discarded altogether.
To improve the compression ratio, the quantized blocks are arranged into a zig-zag order and then coded by the run-length encoding (RLE) algorithm.
Finally, the resulting data for all 8x8 blocks are further compressed with a lossless algorithm, a variant of Huffman encoding. For decompression, inverse transforms of the same steps are applied in reverse order. If the DCT computation is performed with sufficiently high precision, quantization and subsampling are the only lossy operations whereas the others are lossless, so they are reversible.

\section{Related work}
\label{sec:related}
The processing of JPEG compressed data has been widely-explored by many conventional image processing techniques as an alternative to speed up the computation performance in a variety of applications, such as face recognition~\cite{IVC_2009_Delac}, image indexing and retrieval~\cite{MCM_2013_Poursistani}, and many others. 
In deep learning era, the potential of the JPEG compressed domain for neural networks has received limited attention and a few works have emerged in the literature only recently.

To accelerate the training and inference speed, Ehrlich~and~Davis~\cite{ICCV_2019_Ehrlich} reformulated the ResNet architecture to perform its operations directly on the JPEG compressed domain. 
Since the lossless operations used by the JPEG compression algorithm are linear, they can be composed along with other linear operations and then applied to the network weights. 
In this way, the basic operations used in the ResNet architecture, like convolution, batch normalization, etc, were adapted to operate in the JPEG compressed domain. 
For the ReLU activation, which is non-linear, an approximation function was developed.

In a different direction, Deguerre~et~al.~\cite{ITSC_2019_Deguerre} proposed a fast object detection method which takes advantage of the JPEG compressed domain.
For this, the Single Shot multibox Detector (SSD)~\cite{ECCV_2016_Liu} architecture was adapted to accommodate block-wise DCT coefficients as input. 
To preserve the spatial information of the original image, the first three blocks of the SSD network were replaced by a convolutional layer with a filter size of 8x8 and a stride of 8.
In this way, each 8x8 block from JPEG compressed data is mapped into a single position in the feature maps used as input for the next layer. 

Similarly, the neural network introduced by Gueguen~et~al.~\cite{NIPS_2018_Gueguen} is a modified version of the ResNet-50~\cite{CVPR_2016_He} capable of operating directly on DCT coefficients rather than RGB pixels.
After the DCT coefficients are obtained by partial decoding JPEG compressed data, their Cb and Cr components are up-sampled to match the resolution of the Y component.
Next, the Y, Cb, and Cr components are concatenated channel-wise, passed through a batch normalization layer, and fed to the convolution block of the second stage of the ResNet-50. 
Due to the smaller spatial resolution of the DCT inputs, the strides of the early blocks of the second stage were decreased, mimicking the increase in size of the receptive fields in the original ResNet-50.
Also, the second and third stages were changed to accommodate the amount of input channels and to ensure that, at their end, they have the same number of output channels as the original ResNet-50.

However, these changes led to a significant increase in the computational complexity of the ResNet-50 network.
To alleviate its network computation costs and number of parameters, Santos~et~al.~\cite{ICIP_2020_Santos} extended the modified ResNet-50 network of Gueguen~et~al.~\cite{NIPS_2018_Gueguen} to include a Frequency Band Selection (FBS) technique for selecting the most relevant DCT coefficients before feeding them to the network.
The FBS technique relies on the idea that higher frequency information have little visual effect on the image, retaining the lowest frequency coefficients.
Although this approach is efficient, image details are completely lost by discarding high frequency information, which may drop the model accuracy.

\section{Speeding up CNN models designed for DCT input}
\label{sec:approach}
In this paper, we extend the work of Santos~et~al.~\cite{ICIP_2020_Santos}, investigating smarter strategies to reduce the computational complexity and number of parameters of the ResNet-50 network proposed by Gueguen~et~al.~\cite{NIPS_2018_Gueguen}.
In Section~\ref{sec:channel}, we examine learning strategies to reduce the number of channels in the early stages of the ResNet-50 network but without sacrificing any information provided by the DCT coefficients.
In a different direction, Section~\ref{sec:layers} investigates the possibility of reducing the computational complexity by decreasing the number of layers of the network, while attempting to keep the model accuracy.

\subsection{Reducing the number of channels}
\label{sec:channel}
To start, we evaluate the simple idea of reducing the number of channels in the early stages of the modified ResNet-50 of Gueguen~et~al.~\cite{NIPS_2018_Gueguen}, which has been proven to be effective in reducing the computational costs of the network~\cite{ICIP_2020_Santos}.

First, we reduce the number of input channels of the second stage to 64 but we kept the decreased strides at its early blocks, as proposed by Gueguen~et~al.~\cite{NIPS_2018_Gueguen}. 
To accommodate this amount of input channels, we change number of output channels of the second and third stages are changed to be the same as the original ResNet-50.
Then, we evaluate different strategies to reduce the number of DCT inputs from 192 (i.e., 3 color components $\times$ 64 DCT coefficients) to 64. 

Unlike the FBS of Santos~et~al.~\cite{ICIP_2020_Santos}, where the DCT inputs are discarded by hand potentially losing image information, we take advantage of all DCT inputs and learn how to combine them in a data-driven fashion.
For this, we evaluate three different approaches: (1) a linear projection~(LP) of the DCT inputs (Section~\ref{sec:lp}), (2) a local attention~(LA) mechanism (Section~\ref{sec:la}), and (3) a cross channel parametric pooling~(CCPP) (Section~\ref{sec:nin}).

\subsubsection{Linear projection (LP)}
\label{sec:lp}
The ResNet-50 network~\cite{CVPR_2016_He} have residual learning applied to every block of few stacked layers, given by Equation~\ref{eq:lp1}, where $F()$ is the residual mapping to be learned by the $i$-th block of stacked layers, $W_i$ are its parameters, $x$ are the input data, and $y$ are the output feature maps.
\begin{equation}
    \label{eq:lp1}
    y = F ( x, {W_i} ) + x 
\end{equation}

The $F() + x$ operation is executed by a shortcut connection and a element-wise addition, but their dimensions must be equal. When they are not, a $W_s$ linear projection can be applied in order to match the dimension. 
As can be seen in Equation~\ref{eq:lp2}, assuming that $x$ have $n$ input features maps and $W_s$ is a weight matrix of size $m \times n$, the product $W_s \cdot x$ will output in $m$ feature maps, where each one is a linear combination of all the $n$ inputs from $x$.
\begin{equation}
    \label{eq:lp2}
    y = F( x, {W_i} ) + W_s \cdot x
\end{equation}

We apply this linear projection to reduce the number of channels from 192 of the DCT inputs to 64 of the convolution block of the second stage. 
In this way, we consider the DCT inputs as a whole regardless the importance of each of their frequencies to the image content. 
Also, the skewness or kurtosis (shape) of their distribution is preserved by the linear transformation.

\subsubsection{Local attention (LA)}
\label{sec:la}
The local attention proposed by Luong~et al.~\cite{luong2015effective} is a soft attention mechanism used on the machine translation task to analyze a word with a small context window of adjacent words, learning attention maps which focus on relevant parts of the input information.

We adapt this mechanism to be used in the DCT inputs in order to reduce the number of channels from 192 to 64. This is performed according to Equation~\ref{eq:la}, where $x$ is an input with $n$ features maps, $r$ is its reshaped version partitioning it into $m$ groups of $\frac{n}{m}$ channels, $W$ is a weight matrix of size $m \times (\frac{n}{m})$, $y$ is an output with $m$ feature maps, and $\odot$ is the Hadamard product.
\begin{align}
    \label{eq:la}
    r &= reshape\left( x, \left[ m, \frac{n}{m} \right] \right) \\
    s &= W \odot r \nonumber \\
    a_i &= softmax( s_i ), \forall i \in \{1 \dots m\} \nonumber \\
    y_i &= a_{i} \cdot r_{i}, \forall i \in \{1 \dots m\} \nonumber
\end{align}

First, the input $x$ is split into $m$ partitions $r = \{r_1, \dots, r_m\}$ with $\frac{n}{m}$ features maps.
Then, alignment scores $s$ are obtained by computing the Hadamard product between $W$ and $r$. 
For each partition $i \in \{1 \dots m\}$, alignment scores $s_i$ are normalized by applying the softmax function, producing attention maps $a_i$ which are used to amplify or attenuate the focus of the distribution of the input data $r_i$. 
Therefore, the feature map $y_i$ outputted for the $i$-th partition is a linear combination of adjacent channels. 
In this way, we preserve information of the DCT spectrum for the entire range of frequencies.

\subsubsection{Cross Channel Parametric Pooling (CCPP)}
\label{sec:nin}
In a cross channel parametric pooling layer, a weighted linear recombination of the input features maps is performed and then passed though a rectifier linear unit (ReLu)~\cite{lin2013network}.
Min~Lin~et~al.~\cite{lin2013network} proposed to use a cascade of such layers to replace the usual convolution layer of a CNN, since they have enhanced local modeling and the capability of being stacked over each other. 
Formally, a cascaded cross channel parametric pooling is performed according to Equation~\ref{eq:nin}~\cite{lin2013network}, where $f_{i,j,k}^l$ stands for the output of the $l$-th layer, $x_{i,j}$ is the input patch centered at the pixel $(i, j)$, $k$ is used to index the feature maps, $W_{l,k}$ and $b_{l,k}$ are, respectively, weights and biases of the $l$-th layer for the $k$-th filter, and $N$ is the number of layers.     
\begin{align}
    \label{eq:nin}
    f_{i,j,k}^1 &= max\left( 0, W_{1,k}^T \cdot x_{i,j} + b_{1,k} \right) \\
    &\vdots \nonumber \\
    f_{i,j,k}^N &= max\left( 0, W_{N,k}^T \cdot f_{i,j}^{N-1} + b_{N,k} \right)\nonumber
\end{align}

The cross channel parametric pooling is equivalent to a convolutional layer with a kernel size of $1\times1$~\cite{lin2013network}, which is also know as a pointwise convolution~\cite{chollet2017xception}, being capable of projecting the input feature maps into a new channel space, increasing or decreasing the amount of channels.

We used a cross channel parametric pooling layer to reduce the number of channels from 192 to 64.
Similar to the linear projection, the individual importance of each DCT coefficient for the image content is also not taken into account.
On the other hand, the non-linear properties of the ReLU activation encourages the model to learn sparse feature maps, making it less prone to overfitting.

\subsection{Reducing the number of layers}
\label{sec:layers}
The modified version of the ResNet-50 introduced by Gueguen~et~al.~\cite{NIPS_2018_Gueguen} skips first stage of the original ResNet-50, feeding the DCT coefficients to the second stage, which is modified to accommodate the amount of input channels.
In order to reduce the complexity of the network even further, we analyze the effects of skipping the second, third, and fourth stages of the original ResNet-50, but maintaining the stride reduction proposed by Gueguen~et~al.~\cite{NIPS_2018_Gueguen} at the early blocks of the initial stage in which the DCT coefficients are provided as input.

Different from Gueguen~et~al.~\cite{NIPS_2018_Gueguen}, we do not increase the number of input channels at the initial stages, since it would lead to a great increase on the computational complexity of the network.
Instead, we keep them the same as the original ResNet-50, whose the number of input channels in the second, third, fourth, and fifth stages are 64, 128, 256, and 512, respectively.

To accommodate such amount of channels, the strategies presented in the previous section were used to decrease or increase the DCT inputs from 192 (i.e., 3 color components $\times$ 64 DCT coefficients) to the amount of input channels of the initial stage in which they are provided as input.
Notice that the number of DCT coefficients is close to the number of input channels of the third stage, requiring a less drastic reduction than the one needed to feed them on the second stage.
On the other hand, the expected inputs for the fourth and fifth stages have a greater amount of channels than the DCT inputs and, for this reason, they need to be scaled up, however preserving the salient features as the original data.

\section{Experiments and results}
\label{sec:experiments}
Experiments were conducted on the ILSVRC12~\cite{IJCV_2015_Russakovsky} dataset, commonly known as ImageNet, and on a subset of it used by Santos~et~al.~\cite{ICIP_2020_Santos}.
The ImageNet dataset has 1000 classes and is divided into a training set of 1,281,167 images and a test set of 50,000 images.
The ImageNet subset has 211 of the 1000 classes, totaling 268,773 images that were split into a training set with 215,018 images and a test set of 53,755 images.
Image classification tasks at two difficulty levels were considered for this subset: in the coarse granularity, the 211 classes were grouped according to their semantics into 12 categories, namely: ball, bear, bike, bird, bottle, cat, dog, fish, fruit, turtle, vehicle and sign; whereas in the fine granularity, all the 211 classes were used.

All the images were resized to 256 pixels on their shortest side and the crop size for all experiments was $224 \times 224$.
In the experiments, the evaluated networks were trained for 120 epochs with batch size of 128, initial leaning rate of 0.05 reduced by a factor of 10 every 30 epochs, and momentum of 0.9.
Data augmentation with random crop and horizontal flipping was applied on training phase, while on test, only center crop was used.

The experiments were implemented in PyTorch (version 1.2.0) and performed on a machine equipped with two 10-core Intel Xeon E5-2630v4 2.2 GHz processors, 64 GBytes of DDR4-memory, and 1 NVIDIA Titan Xp GPU.
The machine runs Linux Mint 18.1 (kernel 4.4.0) and the ext4 file system.

Section~\ref{sec:results_channels} compares the performance of the different strategies used to reduce the number of channels from 192 to 64 before feeding them to the network.
Section~\ref{sec:results_layers} shows the effects of reducing the number of layers of the network.

\subsection{Effects of reducing the number of channels}
\label{sec:results_channels}
Table~\ref{tab:results_channels} presents a comparison of the computational costs and the accuracy for the coarse and fine granularity task for the ImageNet subset, and for the entire ImageNet.
The computation complexity was measured by the amount of floating point operations (FLOPs) required for passing one image already loaded into the memory through the network and by its number of parameters.
The value inside parentheses is the number of input channels at the initial stage of each network.
\begin{table}[!htb]
    \small
    \centering
    \caption{Comparison of computational complexity (GFLOPS), number of parameters, and accuracy for the original ResNet-50 with RGB inputs and networks using DCT with different strategies to reduce number of input channels.}
    \begin{tabular}{c|c|c|c|c|c}
        \hline
        \hline
        \multirow{3}{*}{\textbf{Approach}}  & \multicolumn{2}{|c|}{\textbf{Computational Cost}} & \multicolumn{3}{|c}{\textbf{Accuracy}}\\
        \cline{2-6}
        & \multirow{2}{*}{\textbf{GFLOPs}} & \multirow{2}{*}{\textbf{Params}} & \multicolumn{2}{|c|}{\textbf{ImageNet Subset}} & \multirow{2}{*}{\textbf{ImageNet}}\\
        \cline{4-5}
        & & & Fine & Coarse &\\
        \hline
        RGB~(3x1)~\cite{CVPR_2016_He}            & 3.86 & 25.6M & 76.28 & 96.49 & 73.46\\
        DCT~(3x64)~\cite{NIPS_2018_Gueguen}      & 5.40 & 28.4M & 70.28 & 94.15 & 72.33\\
        DCT + FBS~(3x32)~\cite{ICIP_2020_Santos} & 3.68 & 26.2M & 69.79 & 94.53 & 70.22\\
        DCT + FBS~(3x16)~\cite{ICIP_2020_Santos} & 3.18 & 25.6M & 68.12 & 93.92 & 67.03\\
        \hline
        DCT + LP~(1x64)                          & 3.20 & 25.6M & 70.08 & 93.17 & 69.62\\
        DCT + LA~(1x64)                          & 3.20 & 25.6M & 69.15 & 94.23 & 69.96\\
        DCT + CCPP~(1x64)                        & 3.20 & 25.6M & 70.09 & 94.85 & 69.73\\
        \hline
        \hline
    \end{tabular}
    \label{tab:results_channels}
\end{table}

For both tasks on the ImageNet subset, the RGB-based network performed better than the DCT-based ones. 
In the fine-grained task, the network of Gueguen et~al.~\cite{NIPS_2018_Gueguen} achieved the highest accuracy among the DCT-based networks, but also have the highest number of parameters and computational complexity, even compared to the RGB-based network.
Similar results were obtained by the networks of Santos~et~al~\cite{ICIP_2020_Santos} and ours in terms of accuracy however greatly reducing the computational complexity and number of parameters.
In the coarse-grained task, our LA and CCPP networks yielded better results than that of Gueguen~et~al.~\cite{NIPS_2018_Gueguen} and a similar performance to the DCT + FBS~(3x32) of Santos~et~al.~\cite{ICIP_2020_Santos}.

On the full ImageNet dataset, the network of Gueguen~et~al.~\cite{NIPS_2018_Gueguen} also achieved the highest accuracy, whereas the results obtained for those of Santos~et~al.~\cite{ICIP_2020_Santos} and ours were similar, with the advantage of reducing the network computation complexity.
Among the strategies we proposed, LA performed slightly better than LP and CCPP.
Compared to the networks of Santos~et~al.~\cite{ICIP_2020_Santos}, our strategies yielded a similar accuracy to DCT + FBS~(3x32), while having a computational complexity similar to DCT + FBS~(3x16), showing that the use of smarter strategies to learn how to reduce the number of input channels is promising.

The computational complexity and number of parameters of all our strategies (LP, LA, and CCPP) are identical and better than the original ResNet-50, proving to be effective for accelerating computation without sacrificing accuracy.

\subsection{Effects of reducing the number of layers}
\label{sec:results_layers}
When stages of the network are skipped, we need to decrease or increase the DCT inputs in order to match the amount of channels expected at the initial stage in which they are provided as input.
For this, we use the CCPP method, since the results for all the strategies presented in Section~\ref{sec:channel} were similar.
This strategy was chosen because it is commonly applied in CNNs in order to obtain channel-wise projections of the feature maps, like in depthwise separable convolutions~\cite{chollet2017xception}.
Table~\ref{tab:complexity} presents the computational complexity and number of parameters of our ResNet-50 using DCT as input when skipping different stages.
\begin{table}[!htb]
    \small
    \centering
    \caption{Comparison of computational complexity (GFLOPS) and number of parameters for our ResNet-50 using DCT as input when skipping different stages and using the CCPP to accommodate the amount of channels expected at the initial stage.}
    \begin{tabular}{l|c|c}
        \hline
        \hline
        \textbf{Approach} & \textbf{GFLOPs} & \textbf{Params} \\
        \hline
        Skip the first stage                             &  3.20 & 25.6M \\
        Skip the first and second stages                 &  2.86 & 25.1M \\
        Skip the first, second, and third stages         &  8.26 & 23.9M \\
        Skip the first, second, third, and fourth stages & 10.76 & 15.8M \\
        \hline
        \hline
    \end{tabular}
    \label{tab:complexity}
\end{table}

As it can be seen, skipping the first and second stages was beneficial, reducing the computational complexity and number of parameters of the network.
However, as more stages were skipped, although the number of parameters is decreased, the computational complexity is greatly increased.
This is due to the decreased strides at the early blocks of the initial stage.
For this reason, the skipping of the first and second stages is the only setting considered in the next experiments, since only it saves the computational costs of the network.   

Table~\ref{tab:results} compares the computational complexity, number of parameters, and accuracy between state-of-the-art methods and our proposed strategy, which skips the first and second stages and uses CCPP to accommodate DCT inputs.
Skipping the first and second stages benefited not only computational costs of the network, but also its accuracy.
In both tasks on the ImageNet subset, this modification led to the best performance among the DCT-based networks.
On the full ImageNet dataset, it achieved the second best accuracy among the DCT-based networks, behind only the modified ResNet-50 of Gueguen~et~al.~\cite{NIPS_2018_Gueguen}, whose computational complexity and number of parameters are considerable bigger.
\begin{table}[!htb]
    \small
    \centering
    \caption{Comparison of computational complexity (GFLOPS), number of parameters, and accuracy for the original ResNet-50 with RGB, state-of-the-art networks designed for DCT, and our strategy for reducing the number of input channels and layers.}
    \begin{tabular}{c|c|c|c}
        \hline
        \hline
        \multirow{2}{*}{\textbf{Approach}} & \multicolumn{2}{|c|}{\textbf{ImageNet Subset}} & \multirow{2}{*}{\textbf{ImageNet}}\\
        \cline{2-3}
        & Fine & Coarse &\\
        \hline
        RGB~(3x1)~\cite{CVPR_2016_He}                  & 76.28 & 96.49 & 73.46\\
        DCT~(3x64)~\cite{NIPS_2018_Gueguen}             & 70.28 & 94.15 & 72.33\\
        DCT + FBS~(3x32)~\cite{ICIP_2020_Santos} & 69.79 & 94.53 & 70.22\\
        DCT + FBS~(3x16)~\cite{ICIP_2020_Santos} & 68.12 & 93.92 & 67.03\\
        \hline
        DCT + CCPP~(1x64)                        & 70.09 & 94.85 & 69.73\\
        DCT + CCPP + skipping 1$^{\mathrm{st}}$ and 2$^{\mathrm{nd}}$ stages~(1x128)  & 71.21 & 94.84 & 70.49\\
        \hline
        \hline
    \end{tabular}
    \label{tab:results}
    \vspace{-4mm}
\end{table}

\section{Conclusion}
\label{sec:conclusions}
In this paper, we addressed the efficiency issues of CNNs designed for the JPEG compressed domain. 
More specifically, we speeded-up a modified version of the ResNet-50 proposed by Gueguen~et~al.~\cite{NIPS_2018_Gueguen} and improved by Santos~et~al.~\cite{ICIP_2020_Santos}.
Although these proposals are effective, they introduced changes in the ResNet-50~\cite{CVPR_2016_He} that either raised the computational costs or lost relevant information from the input.
In contrast, we explored smart strategies to reduce the computational complexity without discarding useful information.

We conducted experiments on the ImageNet dataset, both in a subset and in the whole.
Our results on both datasets showed that learning how to combine all DCT inputs in a data-driven fashion performs better than the FBS technique of Santos~et~al.~\cite{ICIP_2020_Santos}, where the DCT inputs are discarded by hand.
Also, we found that skipping some stages of the network is beneficial, proving to be effective for reducing the computational complexity while retaining accuracy.

As future work, we intend to evaluate other learning strategies for accelerating computation without sacrificing accuracy.
Also, we want to evaluate the use of our strategies with other network architectures, like EfficientNet~\cite{ICML_2019_Tan} and MobileNet~\cite{howard2017mobilenets}.
In addition, we also plan to extend the ideas applied on networks designed for JPEG images to those devised for MPEG videos~\cite{SIBGRAPI_2019_Santos,SIBGRAPI_2020_Santos}.

\iffinalversion
\section*{Acknowledgment}
This research was supported by the FAPESP-Microsoft Research Virtual Institute (grant~2017/25908-6) and the Brazilian National Council for Scientific and Technological Development - CNPq (grant~314868/2020-8). 
\fi


\end{document}